\documentclass{article}
\usepackage{iclr2027/iclr2027_conference,times}
\iclrfinalcopy
\makeatletter
\let\f@nch@gbl\global
\makeatother
\usepackage{amsmath,amsfonts,bm}

\everydisplay{\small}

\def\1{\bm{1}}

\DeclareMathAlphabet{\mathsfit}{\encodingdefault}{\sfdefault}{m}{sl}
\SetMathAlphabet{\mathsfit}{bold}{\encodingdefault}{\sfdefault}{bx}{n}

\usepackage{xcolor}
\definecolor{cvprblue}{rgb}{0.21,0.49,0.749}
\usepackage[colorlinks=true,linkcolor=cvprblue,citecolor=cvprblue,urlcolor=cvprblue]{hyperref}
\usepackage{url}
\usepackage{graphicx}
\usepackage{wrapfig}
\usepackage{needspace}
\usepackage{booktabs}
\usepackage{enumitem}
\usepackage{tabularx}
\usepackage{multirow}
\usepackage{colortbl}
\usepackage{longtable}
\usepackage{amssymb}
\usepackage{etoolbox}
\makeatletter
\patchcmd{\@maketitle}{\lhead{Published as a conference paper at ICLR 2027}}{\lhead{}}{}{}
\patchcmd{\@maketitle}{\vskip 0.3in minus 0.1in}{\vskip 0.05in}{}{}
\patchcmd{\endabstract}{\vskip 1ex}{\vskip -12pt\vskip 0pt}{}{}
\makeatother
\usepackage[skins,breakable,raster]{tcolorbox}

\definecolor{promptblue}{HTML}{000000}
\definecolor{promptback}{HTML}{F5F8FB}
\newtcolorbox{promptbox}[1]{
  enhanced,
  breakable,
  colback=promptback,
  colframe=promptblue!55,
  colbacktitle=promptblue,
  coltitle=white,
  fonttitle=\small\bfseries,
  title={#1},
  boxrule=0.45pt,
  arc=1.5mm,
  left=2.5mm,
  right=2.5mm,
  top=1.5mm,
  bottom=1.5mm,
  before skip=5pt,
  after skip=5pt
}
\newtcolorbox{promptcard}[1]{
  enhanced,
  colback=promptback,
  colframe=promptblue!55,
  colbacktitle=promptblue,
  coltitle=white,
  fonttitle=\footnotesize\bfseries,
  fontupper=\footnotesize,
  title={#1},
  boxrule=0.45pt,
  arc=1mm,
  boxsep=0.5mm,
  left=0.8mm,
  right=0.8mm,
  top=0.5mm,
  bottom=0.5mm,
  toptitle=0.4mm,
  bottomtitle=0.4mm,
  before upper={\setlength{\parskip}{1pt}\setlength{\parindent}{0pt}}
}

\everydisplay{}
\newcommand{\revisionSmallMath}[1]{
    \BeforeBeginEnvironment{#1}{\begingroup\small}
    \AfterEndEnvironment{#1}{\endgroup}
}
\forcsvlist{\revisionSmallMath}{equation,equation*,align,align*,gather,gather*,multline,multline*,displaymath}

\hypersetup{
  pdftitle={Getting Space Right: Agentic Rewards for Image Generation},
  pdfauthor={Xiaofeng Tan, Jiaxin Lin, Haobo Yuan, Tiancheng Shen, Zihan Su, Rui Chen, Hongsong Wang, Ming-Hsuan Yang}
}

\title{Getting Space Right: Agentic Rewards for \\ Image Generation}

\newcommand{\afflogo}[2]{\raisebox{-0.22em}{\includegraphics[height=#1]{#2}}\hspace{0.15em}}

\author{%
\parbox{\dimexpr\textwidth-2\tabcolsep\relax}{\raggedright\parindent0pt
\mbox{Xiaofeng Tan$^{1,2}$\thanks{Work done at Tencent HY. $^{\dagger}$Corresponding author.}} \quad
\mbox{Jiaxin Lin$^{2}$} \quad
\mbox{Haobo Yuan$^{3}$} \quad
\mbox{Tiancheng Shen$^{3}$} \quad
\mbox{Zihan Su$^{4}$} \quad
\mbox{Rui Chen$^{2}$} \quad
\mbox{Hongsong Wang$^{1}$\textsuperscript{$\dagger$}} \quad
\mbox{Ming-Hsuan Yang$^{3}$}\\[0.55em]
\mdseries
\parbox{\linewidth}{\raggedright
\mbox{$^{1}$\afflogo{1.05em}{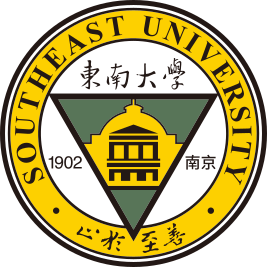}Southeast University}\hspace{0.9em}%
\mbox{$^{2}$\afflogo{1.05em}{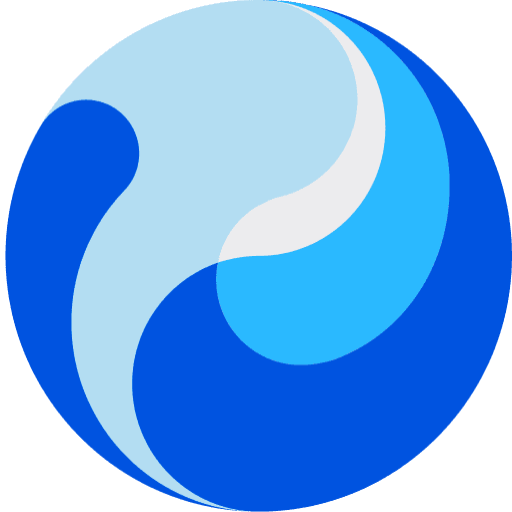}Tencent HY}\hspace{0.9em}%
\mbox{$^{3}$\afflogo{1.05em}{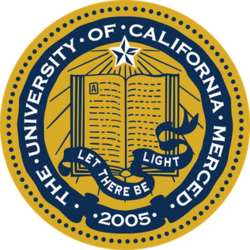}UC Merced}\hspace{0.9em}%
\mbox{$^{4}$\afflogo{1.05em}{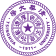}Tsinghua University}}}}

\begin{document}
\maketitle

\begin{abstract}
Reinforcement learning (RL) is promising for spatial text-to-image generation, but existing work largely focuses on policy optimization, assuming accurate rewards and effective online rollouts. We observe in-domain gains with out-of-domain degradation, exposing two limitations: unreliable spatial rewards in open-world settings and rollouts lacking basic-to-complex progression. We propose a curriculum-guided RL framework. First, we introduce \textsc{SpatialCritic}, an agentic reward model equipped with a spatial perception harness that combines reusable verification guidance with specialized vision tools. We then construct spatial preference datasets through targeted counterfactual perturbations and perform agent--harness alignment to improve tool selection and evidence integration efficiently. Second, curriculum learning organizes shared foundational capabilities and their compositional dependencies in a capability graph for mastery-guided rollouts, periodic review, and targeted remediation. On SpaRW-Eval, \textsc{SpatialCritic} achieves 95.1\% verification accuracy, outperforming 320B GLM-5.3-Flash with a $40\times$ smaller 8B backbone and raising average baseline scores from 78.4 to 83.9 on SD3.5-Medium and 83.2 to 86.5 on BAGEL. \mbox{See the \href{https://xiaofeng-tan.github.io/projects/spatialcritic/}{project page}.}
\end{abstract}

\section{Introduction}
\label{sec:intro}

Enhancing spatial intelligent in text-to-image (T2I) generation, i.e., generating accurate spatial relationships faithful to text prompts~\citep{ghosh2023geneval, wang2026everything}, remains a critical and challenging task in T2I post-training. Recently, reinforcement learning (RL) has emerged as a prevailing paradigm for aligning pre-trained models with task-specific goals.

RL aligns a T2I model by generating images from training prompts, evaluating rewards, and updating the policy. Thus, its effectiveness depends on three aspects: (1) rollout organization, (2) reward accuracy, and (3) policy optimization. Existing work mainly advances the third~\citep{black2024training,clark2024directly,liu2025flow,zheng2025diffusionnft,tan2026consistentrft,tan2026pec}, often assuming that rollouts support effective learning and rewards provide reliable feedback. This leaves two questions:
\textit{(Q1)} {Are reward models accurate and robust enough in open-world scenarios?}
\textit{(Q2)} {Can rollout organization support progressive learning from {basic spatial capabilities} to complex spatial compositions?}

\begin{wrapfigure}{r}{0.33\linewidth}
    \centering
    \vspace{-10pt}
    \includegraphics[width=\linewidth]{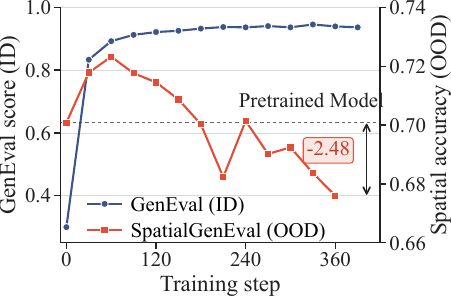}
    \vspace{-18pt}
    \caption{\textbf{Generalization collapse.} ID accuracy improves while OOD performance falls.}
    \label{fig:id_ood_comparison}
    \vspace{-15pt}
\end{wrapfigure}

Unfortunately, these assumptions fail in spatial text-to-image generation: RL-aligned models improve steadily on in-domain (ID) benchmarks but \textit{generalize poorly out of domain (OOD)}, eventually falling below the pre-trained model (Fig.~\ref{fig:id_ood_comparison}). This failure stems from repurposing benchmarks such as GenEval as training prompts and their rule-based scoring pipelines as online rewards, exposing two critical flaws: (1) inaccurate and non-robust rewards, as benchmark-specific scoring generalizes poorly to open-world scenarios, especially under object occlusion~\citep{tang2026enhancing}; and (2) rollouts targeting complex compositions without progressive learning. Reused benchmark prompts emphasize joint constraint satisfaction, neglecting fundamental spatial capabilities and their progressive composition into complex ones.

\begin{figure}[t]
    \centering
    \includegraphics[width=\linewidth]{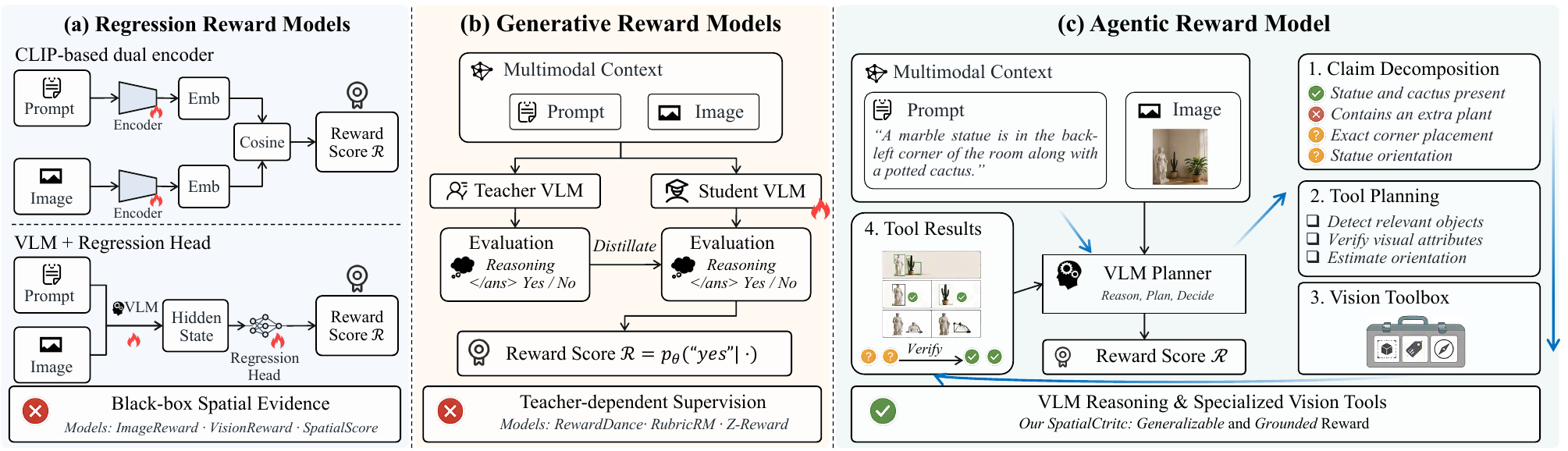}
    \vspace{-12pt}
    \caption{\textbf{Comparison of reward-modeling paradigms.}
    (a) Regression-based scoring; (b) teacher-guided generative evaluation; and (c) our agentic reward model, \textsc{SpatialCritic}.}
    \label{fig:reward_paradigms}
    \vspace{-12pt}
\end{figure}

To overcome the brittleness of rule-based rewards, recent efforts pursue learned reward modeling across two paradigms: \textit{regression-based models} that map dual-encoder or VLM representations into scalar preference scores~\citep{xu2026visionreward, tang2026enhancing}, and \textit{generative models} that emulate deliberative evaluation via reasoning rationales or dynamic rubrics~\citep{wang2026rationalrewards, kan2026rubricrm}, as shown in Fig.~\ref{fig:reward_paradigms}. However, neither ensures reliable spatial verification. Regression models treat foundation models as black-box feature extractors, underutilizing native reasoning and geometric depth priors. Meanwhile, generative models fundamentally rely on ungrounded teacher generation, and struggle to explicitly and reliably resolve geometric quantities such as occlusion and orientation. {Beyond reward modeling, progressive capability learning} remains largely unexplored. \textit{These gaps motivate us to combine a reward model leveraging vision foundation models for verification with a progressive curriculum, jointly ensuring accuracy and generalizability.}

\textbf{Contributions.} We propose a curriculum-guided RL paradigm for spatial T2I post-training, comprising \textsc{SpatialCritic}, an agentic reward model that leverages vision foundation models for robust supervision, and a progressive curriculum from {basic spatial capabilities} to complex compositions.

First, {we design an agentic reward model}, \textsc{SpatialCritic}, that harnesses specialized vision foundation models for grounded spatial verification, optimized via Reinforcement Learning with Verifiable Rewards (RLVR). Its development comprises three components:
\textit{(1) Agent with a spatial perception harness.} To enhance the agent's spatial perception capabilities, we design an agent equipped with a spatial harness that provides reusable verification guidance and executes specialized perceptual tools. Through this harness, the agent decomposes prompts into atomic claims and selectively queries specialized tools under uncertainty.
\textit{(2) Spatial preference dataset construction.} To enhance the agentic ability on the harness, the verifiable RL environment is necessary for post-training. We therefore construct a spatial preference dataset by counterfactually perturbing targeted spatial relations, such as occlusion, depth, and orientation, while keeping other visual content invariant.
\textit{(3) Agent--harness alignment via RLVR.} Building upon the agent and the preference dataset, we optimize the agent's harness ability, such as tool-selection and evidence-integration, using verifiable outcome rewards and token-budget penalties to improve spatial judgment accuracy and efficiency.

Second, {we introduce curriculum learning for spatial T2I post-training}, which organizes policy rollouts from basic spatial capabilities to complex compositions rather than directly optimizing over an unstructured mixture of complex tasks. Our key insight is that \textit{higher-order spatial tasks involve different combinations of shared foundational capabilities}. For example, multi-object layout combines counting, and topology, while occlusion combines depth and topology, sharing topology capabilities. Accordingly, we introduce a {Compositional Capability Graph (CCG)} to capture these dependencies. Building on CCG, capability mastery guides progression, while periodic review identifies compositional weaknesses for targeted remediation. Together, these designs enable feedback-driven progression from basic capabilities to complex compositions.

Finally, \textit{comprehensive experiments} validate accurate preference judgments, effective reward-guided image selection, and improved RL alignment. \textsc{SpatialCritic} scores 95.1\% on SpaRW-Eval. Curriculum RL raises baseline averages to 83.9 on SD3.5-Medium and 86.5 on BAGEL.

\section{Related Work}

\textbf{Reward Models for Visual Generation.}
Visual reward modeling for generation~\citep{su2025safesora, fan2026unidg, su2026generation} has evolved from similarity heuristics~\citep{radford2021learning} to learned human preferences. Early dual encoders~\citep{wu2023hpsv2, zhang2024mps} predict scalar preference scores via Bradley-Terry objectives. To overcome single-scalar bottlenecks, subsequent paradigms deployed multimodal VLMs to predict disentangled rewards across aesthetics, alignment, and visual artifacts~\citep{xu2026visionreward, liu2025improving, ma2025hpsv3, liu2026hpsv3pp, tang2026enhancing}. More recently, generative reward models emit dynamic rubrics, critique rationales, preference tokens, or calibrated score distributions~\citep{wu2025rewarddance, jin2026zreward, kan2026rubricrm, wang2026unified, wang2026rationalrewards}. Closest to our work, SpatialReward~\citep{zhou2026spatialreward} grounds decomposed spatial constraints with expert detectors and reasons over the resulting observations. Its verification procedure is nonetheless determined prior to observation, following a pre-defined routine per task category~\citep{spatialreward2026code}. We instead learn evidence acquisition as a per-claim decision over an open tool catalog, aligned to the harness under outcome supervision.

\textbf{Spatial Intelligent Training Paradigm.}
Recent spatial intelligence studies reveal that learning complex spatial reasoning without grounded perceptual foundations induces severe shortcuts and optimization instability~\citep{li2025spatialladder, liang2026hispatial, yang2026svll}. Consequently, structured spatial data and staged learning are explored in multimodal representation~\citep{cai2025scaling, li2025spatialladder, liang2026hispatial} and embodied planning~\citep{yang2026svll, qian2026xembodied, tan2025sopo, weng2026realign, tan2026easytune, tan2026motionrft}. However, these works primarily target spatial understanding and embodied planning through data scaling, representation adaptation, or staged alignment. In contrast, our work centers on spatial T2I post-training, where a capability dependency graph and an absolute mastery probe guide online prompt scheduling, periodic review, and targeted remediation.

\section{\textsc{SpatialCritic}: Tool-Mediated Agentic Reward Model}
\label{sec:spatialcritic}

\subsection{{Agent with a Spatial Perception Harness}}
\label{sec:spatialcritic_architecture}

\textbf{Problem Formulation.} For spatial intelligence in text-to-image generation, the reward model requires to assess the spatial consistency of the generated image with the spatial constraints in the prompt, including object existence, attribute binding, counting, comparison, 2D position, background placement, distance, depth, orientation, topology, occlusion, multi-object layout, and 3D pose. Action, physical-plausibility, and image-quality checks provide auxiliary context. Preference judgments typically follow either {pairwise comparison} or {pointwise scoring} $S(\mathbf{I},c)$~\citep{xu2023imagereward,wang2025prefgrpo}. Our framework supports both, but primarily uses pairwise comparison. Given $\mathbf{x}=(\mathbf{I}_A,\mathbf{I}_B,c)$, \textsc{SpatialCritic} predicts which image better satisfies the spatial constraints in $c$, yielding $\mathbf{I}_A \succ \mathbf{I}_B$. For score-wise setting, the preference is predicted by $S(\mathbf{I}_A,c)>S(\mathbf{I}_B,c)$.

\textbf{Motivation: From Complex Spatial Relations to Atomic Verification.} Spatial reward assessment benefits from complementary semantic and geometric capabilities. While VLMs excel at open-ended prompt comprehension, they fundamentally struggle with fine-grained spatial relations such as depth ordering and occlusion boundaries~\citep{liu2026depthordering}. Conversely, fixed detector-based evaluators~\citep{ghosh2023geneval} rely on rigid and predefined heuristics and fail on complex open-world spatial relations. For example, consider the prompt \textit{``A palm tree is closer to the camera than a blue beach umbrella''}. The VLM is hard to accurately locate the position of the tree and the umbrella, and estimates their depth ordering, while the detector-based evaluator can provide the precise information. In this case, the complex spatial relation is decomposed into three atomic verification tasks: checking the presence of both objects, the umbrella's color, and their depth ordering.

\textbf{Key Challenges.}
As discussed above, spatial reward modeling benefits from combining semantic reasoning with specialized perceptual evidence. Beyond integrating tool outputs within a predefined verification pipeline~\citep{zhou2026spatialreward}, we seek an agent that dynamically orchestrate vision foundation models for open-world spatial verification. This raises two complementary challenges:
(1)~\textit{for high-frequency and canonical spatial relations}, how to make recurring verification patterns reliable and reusable? (2)~\textit{for complex and unseen spatial relations}, how to adapt these patterns to unfamiliar spatial compositions without being restricted to predefined tools pipeline or relying solely on unverified VLM judgments? Fundamentally, this requires balancing standardized execution on common and frequent relations with compositional flexibility on unseen and novel ones.

\setlength{\intextsep}{0pt}
\begin{wrapfigure}{R}{0.5\textwidth}
    \setlength{\abovecaptionskip}{0pt}
    \setlength{\belowcaptionskip}{0pt}
    \centering
    \vspace{-10pt}
    \includegraphics[width=\linewidth]{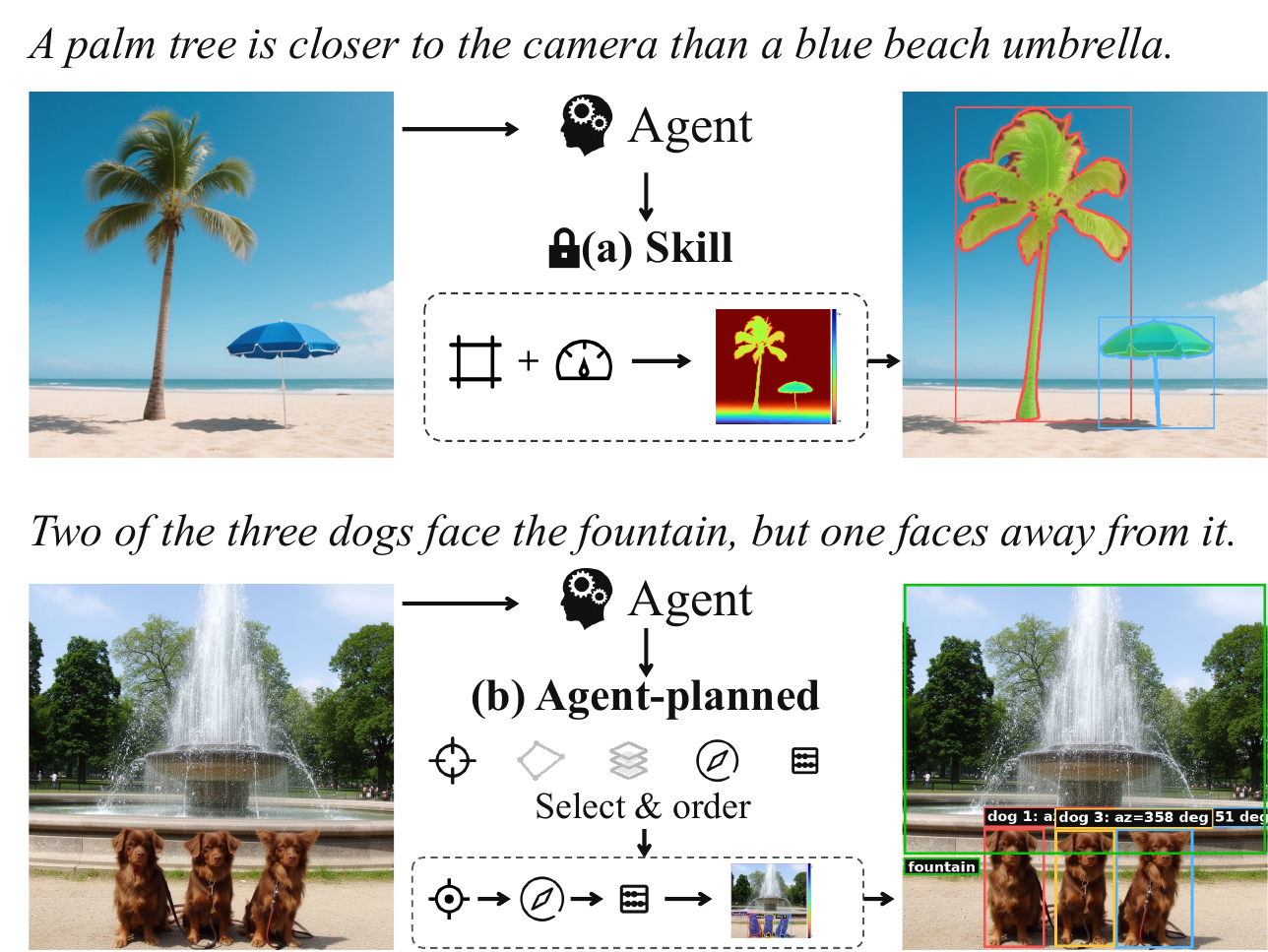}
    \caption{\textbf{Spatial verification.} (a) Fixed mask--depth routine; (b) agent-planned tool use.}
    \label{fig:agent_harness}
\end{wrapfigure}
\textbf{Spatial Perception Harness.}
The harness integrates a modular \textit{perceptual toolkit}, covering grounding, counting, segmentation, depth, 3D geometry, orientation, and contextual checks, with structured \textit{skill guidance}, specifying evidence requirements and output interpretations.
For claims flagged as uncertain during decomposition, the harness guides tool planning along two paths. (1)~When a claim matches a \textit{high-frequency relation}, the agent uses a \textit{standardized skill} with a fixed local tool routine and evidence interpretation, addressing the first challenge. (2)~Otherwise, for \textit{complex or unfamiliar relations} requiring query-specific composition, the harness provides an open tool catalog with capability descriptions, guiding the agent to \textit{compose and sequence targeted tool plans}. In Fig.~\ref{fig:agent_harness}(b), the plan grounds the instances, checks each dog's orientation relative to the fountain, and aggregates these relations to verify the requested counts. This balances robust execution on high-frequency relations with flexible open-world composition. Toolkit details appear in App.~C.

\textbf{Agent Pipeline.}
As shown in Fig.~\ref{fig:reward_paradigms}(c), \textsc{SpatialCritic} verifies spatial consistency through four interactive stages:
(1)~\textit{Claim decomposition}: the VLM policy decomposes the prompt into atomic spatial claims and flags perceptual uncertainties across candidate images;
(2)~\textit{Tool planning}: guided by the spatial perception harness, the agent identifies missing visual evidence and plans targeted tool calls;
(3)~\textit{Tool execution}: the harness invokes the selected vision foundation models and collects structured perceptual observations; and
(4)~\textit{Evidence integration}: the policy integrates these observations into its reasoning to resolve outstanding claims and produce the final preference verdict.

\subsection{Verifiable Spatial Preference Dataset}
\label{sec:dataset_construction}

Although the proposed spatial perception harness balances standardized execution for frequent tasks with flexibility for out-of-distribution cases, efficiency and accuracy remain limited by redundant or failed tool calls, conflicting tool outputs, and ambiguity among similar tools. Fundamentally, these issues reflect a \textit{\textbf{mismatch between the agent's policy and the spatial perception harness}}, requiring the agent to adapt its invocation, selection, and judgment policies to the harness.

This mismatch is difficult to resolve through harness design alone, as agent and tool capabilities vary across tasks and are difficult to align a priori. We therefore employ RLVR to learn adaptive tool selection and evidence integration through interaction, balancing verification accuracy and tool-use cost without predefined execution trajectories. To this end, we construct a spatial preference dataset of image pairs with explicit preference labels for judgment verification and reward computation.

\textbf{Goal.}
To make these preferences effective reward signals for RL, the dataset should satisfy two requirements: \textit{diversity} across spatial tasks and visual contexts, supporting generalizable use of the harness; and \textit{accuracy} of the preference labels, providing reliable feedback for learning tool selection and evidence integration.
We address diversity through \textit{diverse task and scene coverage}, and accuracy through \textit{constraint-targeted preference construction}, which combines explicit prompt constraints with text- and image-level validation.

\textbf{Diverse task and scene coverage.}
Our construction covers a registry of \textit{36 scene settings} and a taxonomy of \textit{13 task categories}, with 3D pose held out from agent training. We balance target-category coverage and sample compatible scenes, supplying their objects, scales, and reference-frame rules to the LLM. Duplicate checks and multiple image generators further broaden textual and visual variation. App.~B.2 lists the task and scene inventories, the complete generation system prompt, and the task-conditioned input fields.

\begin{figure}[t]
    \centering
    \includegraphics[width=\linewidth]{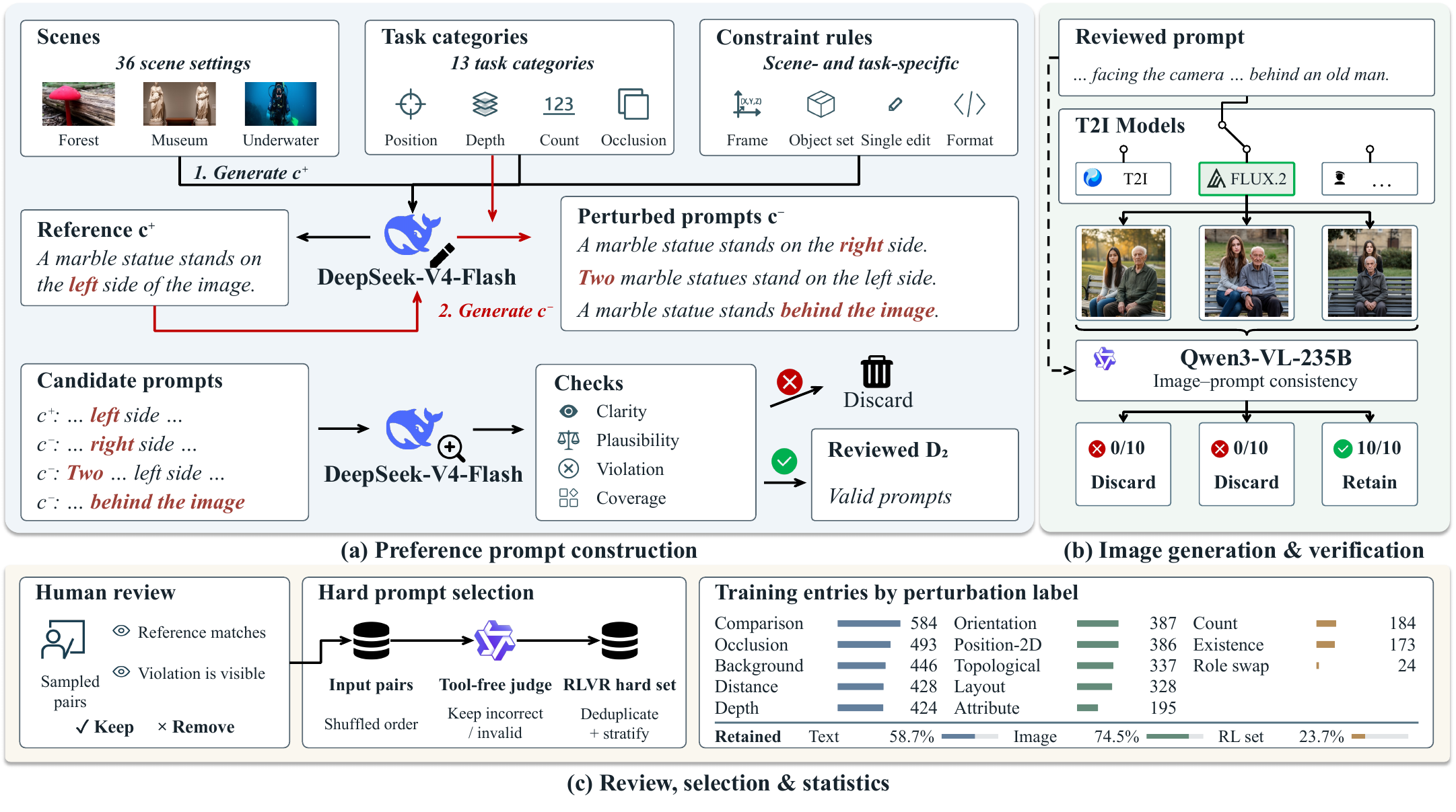}
    \vspace{-18pt}
    \caption{\textbf{Spatial preference dataset construction.} (a) Conditioned prompt generation, perturbation, and review. (b) Multiple image generation and VLM verification. (c) Human review, hard-set selection, and dataset statistics.}
    \vspace{-8pt}
    \label{fig:preference_dataset}
\end{figure}

\textbf{Counterfactual preference construction \& data pipeline.}
To obtain reliable spatial preference labels, we define unambiguous constraints and verify that the generated images form a valid contrast. We construct image preferences from reference prompts $\mathcal{D}_0=\{c\}$ in four stages as shown in Fig.~\ref{fig:preference_dataset}:
(1)~\textit{Prompt-pair construction.} Targeted perturbations provide compositional negatives and counterfactual image pairs~\citep{ma2023crepe,tang2026enhancing}. Conditioned on the target category and scene constraints, DeepSeek-V4-Flash~\citep{deepseekai2026v4} jointly generates an unambiguous reference $c^{+}$ and three alternatives $c^{-}$, forming $\mathcal{D}_1=\{(c^{+},c^{-})\}$ (Fig.~\ref{fig:preference_dataset}(a)). Each alternative is intended to violate one constraint while preserving the others; at least one targets the designated category. The generation instructions are provided in App.~B.2.
(2)~\textit{Text-level validation.} A separate review call to DeepSeek-V4-Flash yields $\mathcal{D}_2=\operatorname{Filter}(\mathcal{D}_1)$ by filtering pairs for clarity, physical plausibility, and explicit violations, assigning perturbation categories, and checking target-category coverage.
(3)~\textit{Image-level verification.} As illustrated in Fig.~\ref{fig:preference_dataset}(b), we render both prompts of each pair in $\mathcal{D}_2$ with FLUX.2-dev~\citep{bfl2025flux2}, Hunyuan-Image-3.0~\citep{tencent2025hunyuanimage3}, and Qwen-Image~\citep{wu2025qwenimage}, using the same generator within a pair, and retain only pairs that pass a Qwen-VL-235B~\citep{bai2025qwen3vl} check of reference compliance and the intended violation, so that preferences reflect image content rather than textual changes alone. We apply best-of-$n$ selection over the rendered candidates and keep the highest-scoring reference-compliant image as $\mathbf{I}^{+}$.
(4)~\textit{Human inspection.} To mitigate automated verification noise, human annotators independently audit candidate pairs retained from image-level verification against identical criteria, confirming that $\mathbf{I}^{+}$ satisfies $c^{+}$ and $\mathbf{I}^{-}$ exhibits the intended violation; failing pairs are removed. The counterfactual prompt $c^{-}$ is discarded after verification, yielding $\mathcal{D}_3=\{(c^{+},\mathbf{I}^{+},\mathbf{I}^{-})\}$, where $\mathbf{I}^{+}\succ\mathbf{I}^{-}$ under $c^{+}$. \mbox{The inspection protocol and quality statistics are in App.~B.1.}

\subsection{Agent--Harness Alignment via RLVR}
\label{sec:spatialcritic_rlvr}

Building on the spatial perception harness and the spatial preference dataset, we align the agent's tool-selection and evidence-integration policy with the harness through RLVR. We optimize the agent policy while keeping the harness and perceptual tools fixed, using verifiable outcome rewards and token-budget penalties to improve spatial judgment accuracy and token efficiency.

\textbf{Hard set selection.}
The initial verification capability of the agent can induce an imbalance between positive and negative outcome rewards. {To provide balanced feedback across spatial tasks, we identify hard where the tool-free base model fails. Combining this difficulty mining with category rebalancing prioritizes complex spatial relations over elementary recognition. Deduplication then yields a balanced training set as shown in Fig.~\ref{fig:preference_dataset}(c), with complete protocols and counts in App.~B.1.} Based on this dataset, we define rewards for preference judgment and tool use.

\textbf{Reward formulation.}
To improve token efficiency while preserving judgment accuracy, we combine an outcome reward with a token-budget penalty conditioned on correct judgments:
\begin{equation}
    \mathcal{R}(\bm{\tau})
    =\mathcal{R}_{\mathrm{out}}(\bm{\tau})
    +\mathbb{I}[\hat{y}(\bm{\tau})=y^{\mathrm{gt}}]\,
    \operatorname{clip}\!\left(\mathcal{R}_{\mathrm{tok}}(\bm{\tau}),-\kappa,0\right).
    \label{eq:reward_decomposition}
\end{equation}
\noindent where $\hat{y}(\bm{\tau})$ is the predicted preference, $y^{\mathrm{gt}}$ is its ground-truth label, and $\kappa$ bounds the auxiliary penalty. We set $\mathcal{R}_{\mathrm{out}}$ to $+1$, $-1$, and $-1.5$ for correct, incorrect, and invalid results, respectively. The token penalty is $\mathcal{R}_{\mathrm{tok}}(\bm{\tau})=-\alpha\max(0,\bm{n}_{\mathrm{tok}}-\bm{n}_0)$, where $\bm{n}_{\mathrm{tok}}$ counts agent-generated tokens, excluding input prompts and tool observations; $\bm{n}_0$ is the free token budget and $\alpha$ penalizes each excess token. This objective encourages accurate judgments, targeted tool use, and concise evidence integration with fewer tokens and lower latency, as illustrated in Figs.~\ref{fig:case_study}, \ref{fig:case_eff}, and~S1.

\textbf{Policy optimization.}
With the dataset and reward, we optimize $\pi_{\bm{\phi}}$ using GRPO~\citep{shao2024deepseekmath}. For each randomly ordered image pair, we sample $G$ rollouts under $\pi_{\bm{\phi}_{\mathrm{old}}}$ following Sec.~\ref{sec:spatialcritic_architecture}, then update the policy using group-relative reward advantages. Details appear in App.~D.1.

\section{Curriculum Learning for Spatial T2I Post-Training}
\label{sec:curriculum}

Building on \textsc{SpatialCritic}, we introduce curriculum learning to address the lack of basic-to-complex progression. This requires {(1) acquiring foundational spatial capabilities; and (2) composing them for combination.} We use a \textit{compositional capability graph} to define these capabilities and their dependencies, and a \textit{progressive learning strategy} to schedule foundational and compositional training accordingly, guided by policy competence, review, and remediation.

\subsection{Compositional Capability Graph}
\label{sec:skill_graph}

\begin{wrapfigure}{r}{0.49\linewidth}
    \vspace{-18pt}
    \centering
    \includegraphics[width=\linewidth]{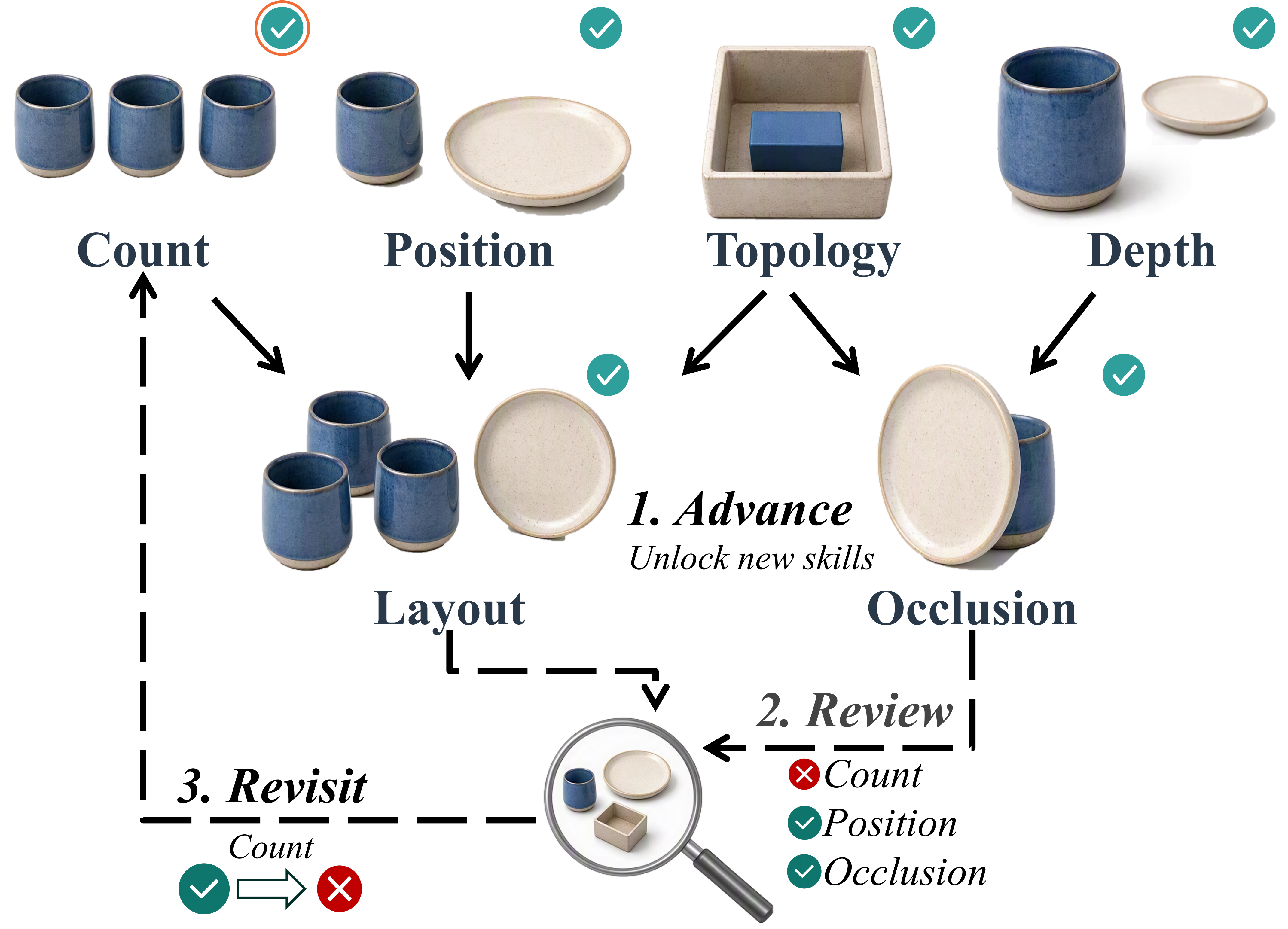}
    \caption{\textbf{CCG-guided progressive learning.} Solid arrows show capability dependencies.}
    \label{fig:skill_curriculum}
\end{wrapfigure}
To guide foundational learning and composition, we examine how spatial capabilities relate. Our key insight is that \textit{higher-order spatial tasks combine shared foundational capabilities in different ways}. CCG represents these foundations and dependencies beyond a single easy-to-hard ordering. Specifically, it is a directed acyclic graph $\mathcal{G}=(\mathcal{V},\mathcal{E})$ with 13 capability nodes, where each edge $(u,v)$ requires $u$ to precede $v$ in training. In Fig.~\ref{fig:skill_curriculum}, Layout and Occlusion share Topology; Layout additionally depends on Count and Position, and Occlusion on Depth. These prescribed dependencies provide an inductive bias for curriculum design, with the complete mapping in App.~D.4. This graph facilitates training scheduling with a progressive learning strategy.

\subsection{Progressive Learning with Review and Remediation}
\label{sec:progressive_learning}

Our progressive learning framework has two components in Fig.~\ref{fig:skill_curriculum}: \textit{(1) forward progression}, which develops foundational capabilities and advances to dependent tasks based on mastery; and \textit{(2) review and remediation}, which addresses forgetting and joint-constraint failures by reassessing learned capabilities individually and jointly.

\begingroup
\textbf{Forward progression.}
Let $\mathcal{P}(v)$ denote the prerequisites of capability $v$ in CCG. At assessment round $t$, we obtain the mean evaluation reward $\mathcal{S}_v^{t}$ for each assessed capability. To reduce sensitivity to outliers, we update its running score as $\bar{\mathcal{S}}_v^{t}=\operatorname{Smooth}(\mathcal{S}_v^{t},\bar{\mathcal{S}}_v^{t-1})$, where $\operatorname{Smooth}$ is an exponential moving average, detailed in App.~D.4. Mastery requires $\bar{\mathcal{S}}_v\ge\tau$ and at least $N_{\min}$ image assessments. To prevent curriculum deadlock, unmastered capabilities permit fallback progression after $N_{\max}$ image assessments. We define the mastered and fallback sets as
\begin{equation}
    \mathcal{M}
    = \underbrace{\bigl\{v\in\mathcal{V}\mid
    \bar{\mathcal{S}}_v\ge\tau,\;
    n_v\ge N_{\min}\bigr\}}_{\text{Mastered capabilities}},
    \quad
    \mathcal{F}
    = \underbrace{\bigl\{v\in\mathcal{V}\mid
    v\notin\mathcal{M},\;
    n_v\ge N_{\max}\bigr\}}_{\text{Unmastered at assessment limit}}.
    \label{eq:mastery_criterion}
\end{equation}
where $n_v$ counts image assessments for capability $v$. $\mathcal{M}$ contains mastered capabilities, while $\mathcal{F}$ contains those still unmastered at the assessment limit. Both sets permit downstream progression, but capabilities in $\mathcal{F}$ remain candidates for review and remediation. A successor capability $v$ becomes \textit{unlocked}, or eligible for training, when all its prerequisites belong to either set:
\begin{equation}
    \operatorname{Unlocked}(v)
    \iff \mathcal{P}(v)\subseteq\mathcal{M}\cup\mathcal{F}.
    \label{eq:gating}
\end{equation}
Fig.~\ref{fig:skill_curriculum} illustrates progression from foundational to composite capabilities. Once Count, Position, Topology, and Depth meet the mastery criterion, they enter $\mathcal{M}$. Layout requires Count, Position, and Topology, while Occlusion requires Topology and Depth. Both sets therefore lie in $\mathcal{M}$, unlocking Layout and Occlusion for training. The \textit{Advance} phase shifts training toward these capabilities, with policy updates on the selected prompts using standard diffusion RL~\citep{liu2025flow}.
\par
\endgroup

\textbf{Review and remediation.}
{Forward progression alone does not ensure capability retention or compositional performance. \textbf{(1)} \textit{Review} periodically evaluates capabilities in $\mathcal{M}\cup\mathcal{F}$ through individual and compositional tasks, updating $\bar{\mathcal{S}}_v^t$ using the same evaluation and smoothing procedure. Capabilities failing the mastery criterion in Eq.~\eqref{eq:mastery_criterion} form the remediation set $\mathcal{C}$ and are removed from $\mathcal{M}$ via $\mathcal{M}\leftarrow\mathcal{M}\setminus\mathcal{C}$. \textbf{(2)} \textit{Remediation} applies targeted training to $\mathcal{C}$ before forward progression resumes. Capabilities satisfying the mastery criterion in either stage enter $\mathcal{M}$ and leave $\mathcal{F}$; those remaining in $\mathcal{F}$ remain eligible for subsequent review and remediation. For example, jointly assessing Count, Position, and Occlusion in Fig.~\ref{fig:skill_curriculum} yields $\mathcal{C}=\{\text{Count}\}$ when only Count fails the mastery criterion. The \textit{Revisit} step then applies counting-specific training to improve performance under joint constraints. App.~D.4 specifies review schedules and training budgets.}

\section{Experiments}
\subsection{Experimental Setups}
\label{sec:experimental_setup}

\textbf{Implementation.}
We fine-tune Qwen3-VL-8B~\citep{bai2025qwen3vl} into \textsc{SpatialCritic} with GRPO on 4,389 spatial preference pairs on 16 GPUs. We also evaluate Qwen3-VL-30B with the same spatial perception harness. We set $\bm{n}_0=400$ free tokens and an excess-token penalty of $\alpha=1.25\times10^{-3}$; see App.~D.2 for reward coefficients. For T2I post-training, we use Flow-GRPO~\citep{liu2025flow} with SD3.5-Medium~\citep{esser2024scaling} and BAGEL~\citep{deng2025emergingpropertiesunifiedmultimodal}. Our curriculum schedules prompts by capability dependencies and mastery; baselines use GenEval~\citep{ghosh2023geneval} or GenEval-2~\citep{kamath2025geneval2}. Each Flow-GRPO iteration samples 24 images for each of 24 prompts, assessing four per group to track mastery with $\tau=0.60$ and $N_{\min}=200$. Further details appear in App.~D.

\textbf{Benchmarks and Datasets.}
We evaluate spatial preference judgments on SpaRW-Eval's 5,572 pairs, reporting per-category and category-size-weighted overall accuracy. Construction follows Sec.~\ref{sec:dataset_construction}, with evaluation details in App.~B.3. We evaluate generation on TIIF-Bench~\citep{wei2025tiifbench}, UniGenBench++~\citep{wang2025unigenbenchpp}, and SpatialGenEval~\citep{wang2026everything}, whose 5,000, 600, and 1,230 prompts cover three instruction-following levels, 10 semantic dimensions, and 10 spatial sub-domains, respectively. The alignment table reports seven benchmark scores and their unweighted mean; ImageReward, PickScore, and HPSv2 separately measure preference quality. Following the benchmark's recommendation to use the Qwen series, we adopt the stronger Qwen3-VL-235B to evaluate generation under the recommended setting.

\textbf{Compared Methods.}
Reward baselines include {Aesthetic Predictor V2.5~\citep{discus0434aestheticv25},} CLIP~\citep{radford2021learning}, ImageReward~\citep{xu2023imagereward}, PickScore~\citep{kirstain2023pick}, HPSv2~\citep{wu2023hpsv2}, HPSv3~\citep{ma2025hpsv3}, UnifiedReward~\citep{wang2025unified}, and SpatialScore~\citep{tang2026enhancing}. We also compare zero-shot VLMs: Qwen2.5-VL~\citep{bai2025qwen25vl}, Qwen3-VL~\citep{bai2025qwen3vl}, Qwen3-Omni~\citep{xu2025qwen3omni}, DeepSeek~\citep{deepseekai2026v41flash}, GLM~\citep{zai2026glm53flash}, Doubao-Seed~\citep{seed2026seed20}, Gemini~\citep{google2026gemini35flash,google2026gemini38flash}, Qwen3.8~\citep{qiu2026qwen38flashnext}, and Grok~\citep{xai2026grok46}. For RL training, we compare three baselines, Flow-GRPO~\citep{liu2025flow}, DiffusionNFT~\citep{zheng2025diffusionnft}, and AWM~\citep{xue2025advantage}, with reward comparisons covering GenEval~\citep{ghosh2023geneval}, GenEval-2~\citep{kamath2025geneval2}, SpatialReward~\citep{zhou2026spatialreward}, and SpatialScore~\citep{tang2026enhancing}.

\begin{table}[t]
    \centering
    \caption{\textbf{Spatial preference accuracy (\%) on SpaRW-Eval.} Per column, \textbf{bold} marks the best open-weight result, \underline{underline} the second best, and \underline{\textbf{\textit{bold italic underline}}} the best across all models. Darker shading indicates higher accuracy. \textbf{Overall} subscripts: relative gain over their backbone VLM (Qwen3-VL-8B/-30B). $^\dagger$: released after July 2026.}
    \label{tab:reward_eval_5k_cat}
    \vspace{0.5em}
    \small
    \setlength{\tabcolsep}{1.25pt}
    \resizebox{\textwidth}{!}{
    \begin{tabular}{l l cccccccccccccc}
        \toprule
        Model & Params & Obj-Attr & Obj-Exist & Count & Bg & Comp & 3D-Pose & Depth & Orient & Pos-2D & Topo & Layout & Occl & Dist & Overall$\uparrow$ \\
        \midrule
        \multicolumn{16}{c}{\textit{Open-weight: reward models}} \\
        \midrule
        Aesthetic predictor & 0.4B & 54.3 & 51.4 & 46.8 & 46.7 & 45.0 & 47.8 & 48.1 & 42.4 & 51.9 & 47.5 & 46.7 & 55.4 & 51.3 & \cellcolor{cvprblue!4}51.6 \\
        CLIP & 0.4B & 72.7 & 85.8 & 84.4 & 52.0 & 66.7 & 76.5 & 52.8 & 62.0 & 50.6 & 63.3 & 66.7 & 63.1 & 59.0 & \cellcolor{cvprblue!4}75.3 \\
        PickScore & 1.0B & 82.3 & 86.4 & 85.3 & 49.3 & 65.0 & 79.3 & 72.2 & 78.3 & 50.6 & 72.5 & 73.3 & 61.5 & 71.8 & \cellcolor{cvprblue!4}81.1 \\
        HPSv2 & 1.0B & 84.3 & 94.0 & 90.8 & 40.0 & 55.0 & 75.4 & 67.6 & 72.8 & 46.8 & 78.3 & 76.0 & 52.3 & 79.5 & \cellcolor{cvprblue!4}83.1 \\
        UnifiedReward & 7B & 85.7 & 91.4 & 89.9 & 61.3 & 73.3 & 87.1 & 63.9 & 71.7 & 57.0 & 84.2 & 81.3 & 66.2 & 89.7 & \cellcolor{cvprblue!6}85.7 \\
        SpatialScore & 7B & 89.1 & \textbf{98.4} & 92.7 & \underline{\textbf{\textit{98.7}}} & 78.3 & 61.7 & 76.9 & 84.8 & \underline{86.1} & 95.0 & 90.7 & 80.0 & \underline{92.3} & \cellcolor{cvprblue!10}87.5 \\
        \midrule
        \multicolumn{16}{c}{\textit{Open-weight: large vision-language models}} \\
        \midrule
        Qwen3-VL-8B & 8B & 89.5 & 82.3 & 85.3 & 58.7 & 70.0 & 91.0 & 58.3 & 67.4 & 48.1 & 82.5 & 81.3 & 67.7 & 79.5 & \cellcolor{cvprblue!4}84.9 \\
        DeepSeek-V4.1-Flash$^\dagger$ & 552B & 92.2 & 94.0 & 95.4 & 70.7 & 88.3 & 91.0 & 70.4 & 79.3 & 74.7 & 91.7 & 86.7 & 72.3 & 76.9 & \cellcolor{cvprblue!19}90.9 \\
        Qwen3-VL-235B & 235B & 95.3 & 94.6 & 93.6 & 88.0 & 81.7 & 95.7 & 68.5 & 76.1 & 67.1 & 95.0 & 90.7 & 70.8 & \underline{92.3} & \cellcolor{cvprblue!25}93.3 \\
        Qwen3-Omni-30B & 35B & 95.3 & 95.5 & 95.4 & 69.3 & 88.3 & 95.1 & 74.1 & 81.5 & 68.4 & 91.7 & 89.3 & 80.0 & 89.7 & \cellcolor{cvprblue!26}93.5 \\
        Qwen3-VL-30B & 30B & 95.3 & 95.9 & 92.7 & 73.3 & 83.3 & 94.4 & 84.3 & 78.3 & 69.6 & \underline{95.8} & 85.3 & 80.0 & \textbf{94.9} & \cellcolor{cvprblue!26}93.7 \\
        GLM-5.3-Flash$^\dagger$ & 320B & 93.9 & \underline{96.7} & \underline{\textbf{\textit{99.1}}} & 77.3 & \underline{93.3} & \textbf{97.0} & 87.0 & \textbf{92.4} & 74.7 & \underline{95.8} & \textbf{94.7} & \textbf{86.2} & \textbf{94.9} & \cellcolor{cvprblue!28}94.5 \\
        \midrule
        \rowcolor{gray!15}\multicolumn{16}{c}{\textit{Agentic Reward Model}} \\
        \midrule
        \rowcolor{gray!15}\textsc{SpatialCritic} & 8B & 94.2 & 93.6 & 90.8 & \underline{90.7} & 88.3 & 92.6 & \underline{88.0} & 82.6 & 79.7 & 93.3 & \underline{93.3} & \underline{81.5} & 89.7 & \cellcolor{cvprblue!24}92.9$_{+9.4\%}$ \\
        \rowcolor{gray!15}\quad w/ AHA & 8B & \underline{96.0} & 96.2 & \underline{98.2} & 84.0 & 88.3 & 95.6 & \textbf{90.7} & 85.9 & 81.0 & 95.0 & \textbf{94.7} & \underline{81.5} & 89.7 & \cellcolor{cvprblue!30}\underline{95.1}$_{+12.0\%}$ \\
        \rowcolor{gray!15}\textsc{SpatialCritic} & 30B & \textbf{96.6} & 96.0 & 95.4 & 89.3 & \textbf{95.0} & \underline{95.9} & \underline{88.0} & \underline{90.2} & \textbf{91.1} & \underline{\textbf{\textit{97.5}}} & 90.7 & 78.5 & \underline{92.3} & \cellcolor{cvprblue!31}\textbf{95.5}$_{+1.9\%}$ \\
        \midrule
        \multicolumn{16}{c}{\textit{Proprietary: large vision-language judges}} \\
        \midrule
        Doubao-Seed-2.0-Lite & --- & 86.0 & 92.3 & 93.6 & 68.0 & 80.0 & 90.6 & 65.7 & 76.1 & 68.4 & 91.7 & 88.0 & 61.5 & 76.9 & \cellcolor{cvprblue!10}87.2 \\
        Gemini 3.5 Flash & --- & 90.3 & 93.1 & 92.7 & 70.7 & 65.0 & 88.3 & 56.5 & 76.1 & 63.3 & 90.0 & 81.3 & 52.3 & 76.9 & \cellcolor{cvprblue!13}88.4 \\
        Qwen3.8-Flash-Next$^\dagger$ & --- & \underline{\textbf{\textit{97.0}}} & 97.9 & \underline{\textbf{\textit{99.1}}} & 84.0 & 81.7 & 96.6 & 76.9 & 94.6 & 79.7 & 96.7 & \underline{\textbf{\textit{97.3}}} & 76.9 & 92.3 & \cellcolor{cvprblue!32}95.9 \\
        Grok 4.6$^\dagger$ & --- & 96.6 & 98.3 & 98.2 & 88.0 & 91.7 & 95.4 & \underline{\textbf{\textit{93.5}}} & 91.3 & 92.4 & 95.8 & 94.7 & 84.6 & 97.4 & \cellcolor{cvprblue!33}96.3 \\
        Doubao-Seed-2.0-Pro & --- & 96.9 & 98.5 & 97.2 & 93.3 & \underline{\textbf{\textit{96.7}}} & 96.6 & 84.3 & \underline{\textbf{\textit{95.7}}} & 83.5 & 95.8 & \underline{\textbf{\textit{97.3}}} & 83.1 & \underline{\textbf{\textit{100.0}}} & \cellcolor{cvprblue!34}96.6 \\
        Gemini 3.8 Flash$^\dagger$ & --- & 96.4 & \underline{\textbf{\textit{98.9}}} & 98.2 & 86.7 & 95.0 & \underline{\textbf{\textit{97.4}}} & 91.7 & 87.0 & \underline{\textbf{\textit{97.5}}} & 95.0 & 94.7 & \underline{\textbf{\textit{87.7}}} & 94.9 & \cellcolor{cvprblue!34}\underline{\textbf{\textit{96.7}}} \\
        \bottomrule
    \end{tabular}
    }
\end{table}

\subsection{Main Results}
\label{sec:main_results}

\textbf{Accuracy of Preference Judgments.}
Tab.~\ref{tab:reward_eval_5k_cat} shows that 8B \textsc{SpatialCritic} achieves 95.1\% verification accuracy, compared with 84.9\% for Qwen3-VL-8B and 87.5\% for SpatialScore. Agent--harness alignment raises accuracy from 92.9\% to 95.1\%, surpassing 320B GLM-5.3-Flash at 94.5\%. The 30B variant reaches 95.5\%, approaching Gemini 3.8 Flash at 96.7\%.

\begin{figure}[t]
    \centering
    \includegraphics[width=\linewidth]{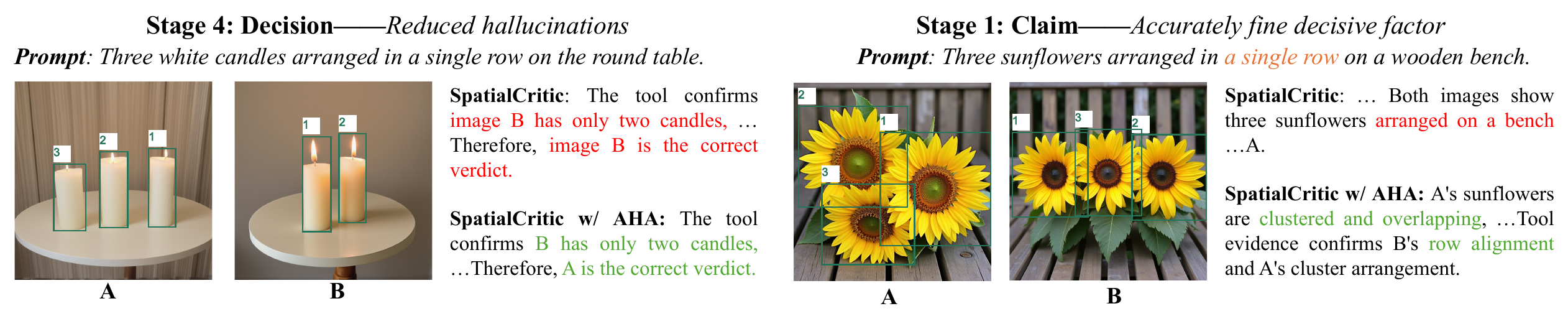}
    \vspace{-20pt}
    \caption{\textbf{Agent--harness alignment cases.} {Left: alignment reduces final-judgment hallucinations, correctly selecting A because evidence shows only two candles in B. Right: it identifies the decisive single-row constraint and distinguishes B's row from A's cluster using tool evidence. App. Fig.~S1 illustrates improved planning.}}
    \vspace{-6pt}
    \label{fig:case_study}
\end{figure}

\textbf{Quality of Reward Signals.}
In Tab.~\ref{tab:main_results}, \textsc{SpatialCritic} leads Best-of-$N$ selection with 74.4\% accuracy, exceeding GLM-5.3-Flash at 71.8\% and the selection-free baseline at 70.8\%. It leads in nine of ten categories, including all five geometry and spatial-relation categories. Position accuracy reaches 73.3\%, versus 67.6\% for the best baseline.

\begin{table}[t]
    \centering
    \caption{\textbf{Reward-guided Best-of-N image generation.} Bold and underlined entries indicate the best and second-best results.}
    \label{tab:main_results}
    \vspace{0.5em}
    \small
    \setlength{\tabcolsep}{7pt}
    \resizebox{\textwidth}{!}{
    \begin{tabular}{l ccccc ccccc c}
        \toprule
        Selector & Attr. & Object & Comp. & Causal & Motion & Layout & Occl. & Orient. & Pos. & Prox. & Overall$\uparrow$ \\
        \midrule
        SD3.5-Medium (base) & 87.89 & 94.63 & 38.62 & 80.57 & 72.68 & 72.68 & 59.35 & 65.77 & 66.59 & 68.86 & 70.76 \\
        \midrule
        Qwen2.5-VL-7B & 88.10 & 94.40 & 38.50 & 80.60 & 72.20 & 72.50 & \underline{59.70} & 65.30 & 65.60 & 68.50 & 70.54 \\
        PickScore & \underline{89.43} & 94.80 & 39.11 & 80.49 & 72.52 & 73.50 & 57.80 & 65.12 & 65.69 & 69.51 & 70.80 \\
        ImageReward & 88.94 & \textbf{95.37} & 38.37 & 81.14 & 73.74 & 72.20 & 58.05 & 66.10 & 65.12 & 69.27 & 70.83 \\
        CLIP & 89.19 & 94.80 & \underline{40.57} & 81.14 & 72.76 & 72.85 & 59.51 & 63.90 & 65.45 & 69.59 & 70.98 \\
        HPSv2 & 89.11 & 94.88 & 39.84 & 80.49 & 73.33 & 72.20 & 58.70 & \underline{67.40} & 65.53 & 69.59 & 71.11 \\
        HPSv3 & 88.94 & 94.07 & 40.24 & 80.81 & 74.23 & 73.33 & 58.05 & 65.85 & 66.42 & \underline{71.22} & 71.32 \\
        UnifiedReward & 88.70 & 94.88 & 39.35 & \underline{81.30} & 74.96 & 72.93 & 58.21 & 66.18 & \underline{67.64} & 70.00 & 71.41 \\
        GLM-5.3-Flash & 89.27 & 94.39 & 39.92 & \underline{81.30} & \underline{75.85} & \underline{74.31} & 57.72 & 66.42 & \underline{67.64} & 70.73 & \underline{71.76} \\
        \midrule
        \rowcolor{gray!15}\textsc{SpatialCritic} & \textbf{90.41} & \underline{94.96} & \textbf{43.17} & \textbf{83.25} & \textbf{79.43} & \textbf{77.72} & \textbf{60.81} & \textbf{68.21} & \textbf{73.33} & \textbf{73.09} & \textbf{74.44} \\
        \bottomrule
    \end{tabular}
    }
    \vspace{-6pt}
\end{table}

\ifdefined\Needspace\Needspace{10\baselineskip}\fi
\begin{wrapfigure}{r}{0.65\linewidth}
    \centering
    \vspace{-6pt}
    \includegraphics[width=\linewidth]{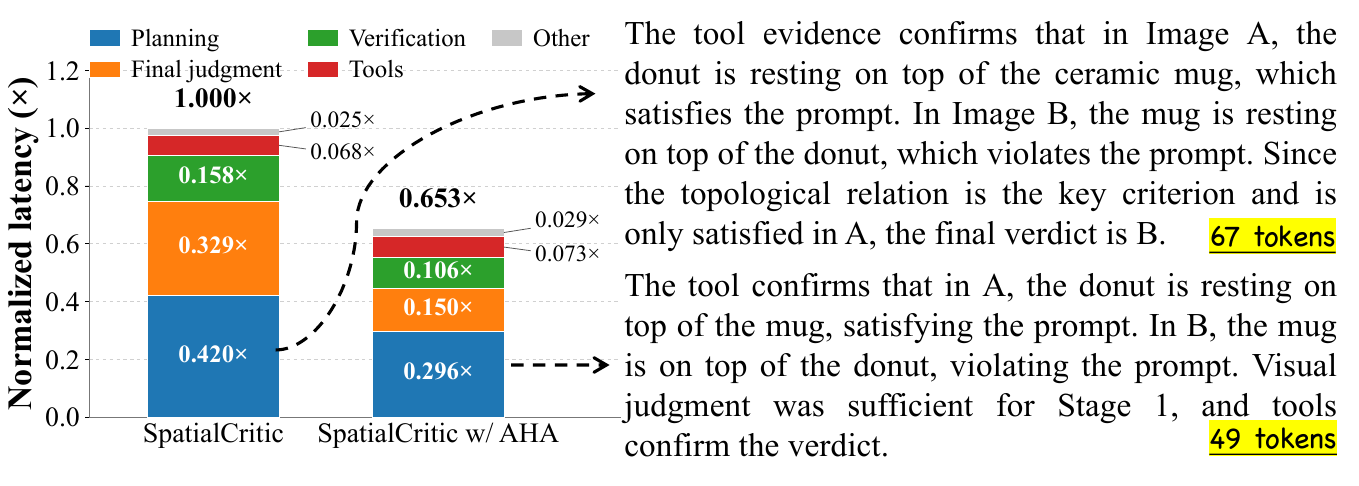}
    \vspace{-20pt}
    \caption{\textbf{Efficiency and decision correction.} Alignment reduces tokens and latency while rectifying judgment errors.}
    \label{fig:case_eff}
    \vspace{2pt}
\end{wrapfigure}
\textbf{Effectiveness of Harness and Agent--Harness Alignment.}
{The harness provides verification guidance and grounded evidence, while alignment improves evidence integration to mitigate hallucinations and support reliable judgments. In Fig.~\ref{fig:case_eff}, alignment rectifies the unaligned verdict error while reducing tokens from 67 to 49 and normalized latency from 1.000 to 0.653, yielding a more concise and accurate judgment. Figs.~\ref{fig:case_study} and~S1 illustrate targeted constraint identification, fewer irrelevant tool calls, and active verification of uncertain claims.}

\textbf{Effectiveness in RL Post-Training.}
Tab.~\ref{tab:gen_align} {shows that SpatialCritic without CL raises averages from 78.4 to 81.9 on SD3.5-Medium and 83.2 to 83.7 on BAGEL, but does not consistently lead. On SD3.5-Medium, its average is below DiffusionNFT + GenEval-2 (82.3), and SpatialGenEval Spatial accuracy (67.0\%) trails SpatialScore (67.3\%) and Flow-GRPO + GenEval-2 (68.4\%). Adding CL raises averages to 83.9 and 86.5 and Spatial accuracy to 69.5\% and 77.8\%, respectively.} The ablation on curriculum dataset in App.~A.3 confirms the effectiveness of curriculum scheduling.

\begin{table}[t]
    \centering
    \caption{\textbf{Reward-guided RL post-training.} Bold marks the best result within each backbone; CL denotes curriculum learning. $^{\dagger}$This non-curriculum setting uses GenEval-2 prompts without its benchmark-specific questions.}
    \label{tab:gen_align}
    \label{tab:gen_align_bagel}
    \label{tab:gen_align_bagel_1of3}
    \vspace{0.5em}
    \small
    \setlength{\tabcolsep}{5pt}
    \resizebox{\textwidth}{!}{
    \begin{tabular}{l l ccc cc cc ccc c}
        \toprule
        \multirow{2}{*}{Method} & \multirow{2}{*}{Reward} & \multicolumn{3}{c}{TIIF-Bench} & \multicolumn{2}{c}{UniGenBench++} & \multicolumn{2}{c}{SpatialGenEval} & \multicolumn{3}{c}{Preference-model scores} & \multirow{2}{*}{Avg} \\
        \cmidrule(lr){3-5} \cmidrule(lr){6-7} \cmidrule(lr){8-9} \cmidrule(lr){10-12}
        & & BR & AR & RR & 2D & 3D & Basic & Spatial & IR & PS & HPS & \\
        \midrule
        SD3.5-Medium (base) & --- & 82.2 & 77.6 & 68.0 & 82.9 & 81.4 & 92.0 & 65.0 & 0.740 & 0.851 & 0.283 & 78.4 \\
        \midrule
        \quad w/ Flow-GRPO & GenEval & 57.8 & 53.7 & 45.4 & 43.6 & 38.7 & 75.9 & 48.1 & -1.317 & 0.727 & 0.162 & 51.9 \\
        \quad w/ DiffusionNFT & GenEval & 68.5 & 57.7 & 56.8 & 66.9 & 44.1 & 65.9 & 53.7 & -1.396 & 0.739 & 0.141 & 59.1 \\
        \quad w/ AWM & GenEval & 72.8 & 69.4 & 62.9 & 69.4 & 44.1 & 83.2 & 56.1 & -1.116 & 0.757 & 0.165 & 65.4 \\
        \quad w/ Flow-GRPO & SpatialReward & 81.3 & 80.1 & 69.6 & 76.2 & 77.3 & 92.6 & 65.0 & 0.451 & 0.823 & 0.259 & 77.4 \\
        \quad w/ Flow-GRPO & SpatialScore & 83.4 & 80.9 & 69.4 & 92.1 & 78.7 & 93.7 & 67.3 & 0.885 & \textbf{0.857} & \textbf{0.306} & 80.8 \\
        \quad w/ Flow-GRPO & GenEval-2 & 84.8 & 81.1 & 72.1 & 84.2 & 84.1 & 93.7 & 68.4 & 0.241 & 0.795 & 0.226 & 81.2 \\
        \quad w/ DiffusionNFT & GenEval-2 & 85.1 & 83.7 & 73.4 & 89.0 & 82.8 & 94.0 & 68.3 & 0.588 & 0.823 & 0.271 & 82.3 \\
        \midrule
        \rowcolor{gray!15}\quad w/ Flow-GRPO$^{\dagger}$ & \textsc{SpatialCritic} & 83.3 & 80.3 & 69.8 & \textbf{93.3} & \textbf{85.3} & 94.4 & 67.0 & 0.857 & 0.855 & 0.304 & 81.9 \\
        \rowcolor{gray!15}\quad w/ Flow-GRPO (CL) & \textsc{SpatialCritic} & \textbf{86.2} & \textbf{83.9} & \textbf{74.1} & \textbf{93.3} & \textbf{85.3} & \textbf{95.0} & \textbf{69.5} & \textbf{0.908} & \textbf{0.857} & 0.305 & \textbf{83.9} \\
        \midrule
        BAGEL (base) & --- & 86.4 & 82.5 & 73.8 & 91.8 & 82.8 & 94.2 & 70.9 & 0.753 & 0.855 & 0.283 & 83.2 \\
        \midrule
        \quad w/ DiffusionNFT & GenEval & 86.8 & 79.5 & 74.5 & 92.3 & 82.4 & 90.8 & 67.8 & \textbf{0.825} & 0.837 & 0.264 & 82.0 \\
        \quad w/ DiffusionNFT & GenEval-2 & 87.6 & 80.7 & \textbf{77.3} & 90.6 & 87.3 & 92.2 & 70.0 & 0.764 & 0.826 & 0.257 & 83.7 \\
        \quad w/ Flow-GRPO & GenEval & 85.0 & 78.4 & 72.4 & 88.0 & 77.3 & 78.9 & 62.3 & 0.666 & 0.832 & 0.268 & 77.5 \\
        \quad w/ Flow-GRPO & GenEval-2 & 86.9 & 80.9 & 74.9 & 90.8 & 84.0 & 80.5 & 65.6 & 0.624 & 0.817 & 0.233 & 80.5 \\
        \midrule
        \rowcolor{gray!15}\quad w/ Flow-GRPO$^{\dagger}$ & \textsc{SpatialCritic} & 86.1 & 82.2 & 74.8 & 93.2 & 82.7 & 93.6 & 73.1 & 0.775 & 0.890 & 0.294 & 83.7 \\
        \rowcolor{gray!15}\quad w/ Flow-GRPO (CL) & \textsc{SpatialCritic} & \textbf{88.3} & \textbf{83.9} & 77.1 & \textbf{94.6} & \textbf{89.3} & \textbf{94.7} & \textbf{77.8} & 0.802 & \textbf{0.913} & \textbf{0.298} & \textbf{86.5} \\
        \bottomrule
    \end{tabular}
    }
    \vspace{-2pt}
\end{table}

\section{Conclusion}

In this work, we investigate spatial T2I post-training from the perspectives of reward accuracy and rollout organization. We identify two limitations beyond policy optimization: unreliable spatial rewards and rollouts that target complex compositions without progressive learning. First, we introduce \textsc{SpatialCritic}, which combines a spatial perception harness with specialized vision tools for grounded preference judgments. Spatial preference data and agent--harness alignment via RLVR improve its tool selection and evidence integration. Second, we introduce curriculum learning based on a Compositional Capability Graph, guiding foundational capability development and subsequent composition through mastery-based progression, review, and remediation. Experiments demonstrate accurate preference judgments on SpaRW-Eval, effective reward signals for Best-of-$N$ selection, and improved RL alignment on SD3.5-Medium and BAGEL.

\textbf{Limitations.}
Our evaluation uses a fixed tool configuration, leaving transfer to updated or alternative tools unexplored. Additionally, robustness to temporary tool failures or missing outputs also remains untested. Future work will investigate cross-tool transfer and failure-aware fallback strategies.

\subsection*{AI use statement}

We used generative models to construct synthetic spatial prompts and images and to review candidate preference labels, as described in Sec.~\ref{sec:dataset_construction} and App.~B.1. We used AI assistants for literature search and summarization to support related-work discovery. We also used AI tools to aid and polish paper writing, and for feedback on experimental analysis and inspection of data-processing code and metadata.

\subsection*{Ethics statement}

This work studies spatial verification and text-to-image generation using synthetic preference data and existing models and benchmarks. Synthetic data can inherit biases from its generators, and model-based verification can introduce systematic annotation errors. Human inspection provides an additional check but does not guarantee unbiased or error-free labels. Improved spatial fidelity may also make misleading synthetic images more convincing; spatial correctness should not be interpreted as evidence of authenticity or safety. Model and dataset reuse and redistribution remain subject to the applicable licenses and access conditions.

\subsection*{Reproducibility statement}

We describe the perceptual toolkit in App.~C, preference-data construction and selection in App.~B.1, and generation prompts in App.~B.2. App.~B.3 details the evaluation data and protocol. Training settings, reward coefficients, policy optimization, and curriculum parameters appear in App.~D, D.2, D.1, and~D.4. We will open-source our code upon acceptance.

\bibliography{references}
\bibliographystyle{iclr2027/iclr2027_conference}

\end{document}